%% file: main.tex
\documentclass[12pt]{article}

\usepackage{sbc-template}

\usepackage{graphicx,url}
\usepackage{booktabs}
\usepackage{dirtytalk}
\usepackage{csquotes}
\usepackage{adjustbox}
\usepackage[utf8]{inputenc}
\usepackage{lscape}
\usepackage{longtable}   % Para tabelas que se estendem por múltiplas páginas
     
\title{Aplicação de Large Language Models na Análise e Síntese de Documentos Jurídicos: Uma Revisão de Literatura}

\author{Matheus Belarmino\inst{1}, Rackel Coelho\inst{1}, Roberto Lotufo\inst{2}, Jayr Pereira\inst{1} }

\address{Universidade Federal do Cariri (UFCA)\\
  Juazeiro do Norte -- CE -- Brasil
\nextinstitute
  NeuralMind.ai \\
  Campinas -- SP -- Brasil
  \email{matheus.pereira@aluno.ufca.edu.br, jayr.pereira@ufca.edu.br}
}

\begin{document} 

\maketitle

\begin{abstract}
Large Language Models (LLMs) have been increasingly used to optimize the analysis and synthesis of legal documents, enabling the automation of tasks such as summarization, classification, and retrieval of legal information. This study aims to conduct a systematic literature review to identify the state of the art in prompt engineering applied to LLMs in the legal context. The results indicate that models such as GPT-4, BERT, Llama 2, and Legal-Pegasus are widely employed in the legal field, and techniques such as Few-shot Learning, Zero-shot Learning, and Chain-of-Thought prompting have proven effective in improving the interpretation of legal texts. However, challenges such as biases in models and hallucinations still hinder their large-scale implementation. It is concluded that, despite the great potential of LLMs for the legal field, there is a need to improve prompt engineering strategies to ensure greater accuracy and reliability in the generated results.
\end{abstract}
     
\begin{resumo} 
Os Modelos de Linguagem de Grande Escala (LLMs) têm sido cada vez mais utilizados para otimizar a análise e síntese de documentos jurídicos, possibilitando a automatização de tarefas como sumarização, classificação e recuperação de informações legais. Este estudo tem como objetivo realizar uma revisão sistemática da literatura para identificar o estado da arte em engenharia de prompts aplicada a LLMs no contexto jurídico. Os resultados indicam que modelos como GPT-4, BERT, Llama 2 e Legal-Pegasus são amplamente empregados na área jurídica, e técnicas como Few-shot Learning, Zero-shot Learning e Chain-of-Thought prompting têm se mostrado eficazes para melhorar a interpretação de textos legais. No entanto, desafios como a presença de vieses nos modelos e a ocorrência de alucinações ainda dificultam sua implementação em larga escala. Conclui-se que, apesar do grande potencial dos LLMs para o direito, há a necessidade de aprimoramento das estratégias de engenharia de prompts para garantir maior precisão e confiabilidade nos resultados gerados. 
\end{resumo}

\section{Introdução}

Os avanços em inteligência artificial e aprendizado de máquina têm revolucionado diversas áreas do conhecimento, incluindo o direito. A crescente digitalização dos processos jurídicos gerou um volume exponencial de documentos legais, tornando cada vez mais desafiador para advogados, juízes e pesquisadores processar, interpretar e extrair informações relevantes de maneira eficiente. Nesse contexto, os modelos de linguagem de grande escala (LLMs, do inglês, Large Language Models) surgem como uma tecnologia promissora para auxiliar na análise e síntese de documentos jurídicos \cite{}.

Os LLMs são modelos de aprendizado de máquina treinados em vastos conjuntos de dados textuais para compreender e gerar linguagem natural. Esses modelos são baseados na arquitetura de Transformers \cite{vaswani2017attention}, que permite que eles aprendam diferentes tarefas a partir de grandes volumes de texto por meio de mecanismos de auto-atenção. Dentre os LLMs mais conhecidos estão o GPT-4, Gemini, LLaMa e Sabiá, um modelo voltado para o português brasileiro \cite{abonizio2025sabia3technicalreport}. Esses modelos são capazes de realizar tarefas como responder perguntas, traduzir textos, resumir documentos, classificar informações e até mesmo gerar conteúdos com coerência contextual.

A aplicação de LLMs no contexto legal tem o potencial de otimizar significativamente processos como revisão contratual, predição de decisões judiciais, recuperação de jurisprudência e geração automática de resumos legais. No entanto, apesar de seus avanços, esses modelos ainda enfrentam desafios como alucinações (geração de informações incorretas), viés nos dados de treinamento e dificuldades na interpretação de contextos jurídicos específicos. Assim, um dos principais focos de pesquisa atualmente é a utilização de técnicas de engenharia de \textit{prompts} para melhorar a precisão e relevância das respostas dos LLMs em contextos jurídicos \cite{peixoto2023gpts}.

A engenharia de \textit{prompts} refere-se ao processo de formular comandos ou instruções otimizadas para obter as melhores respostas possíveis dos LLMs. Técnicas como \textit{Zero-shot Learning} (quando o modelo responde sem exemplos), \textit{Few-shot Learning} (quando é treinado com poucos exemplos) e \textit{Chain-of-Thought} (quando se estimula um raciocínio passo a passo) têm sido amplamente estudadas para aprimorar a performance dos LLMs em tarefas jurídicas complexas. Pesquisadores vêm explorando essas estratégias para reduzir erros e viabilizar o uso confiável desses modelos no setor jurídico \cite{moura2023literacia}.

Apesar da ampla gama de pesquisas sobre LLMs aplicados ao direito, os estudos existentes estão dispersos e não sistematizados em uma única fonte. Isso dificulta a compreensão do estado da arte e a identificação de lacunas e oportunidades de aprimoramento na aplicação dessas tecnologias no contexto jurídico. Atualmente, não há uma revisão consolidada na literatura que reúna e analise criticamente os avanços recentes na engenharia de \textit{prompts} aplicada aos LLMs para documentos jurídicos.
Diante dessa lacuna, o objetivo principal desta pesquisa é realizar uma revisão da literatura para identificar o estado da arte em técnicas de engenharia de \textit{prompts} aplicadas a LLMs na análise e síntese de documentos jurídicos complexos em Língua Portuguesa.

\section{Metodologia}

% Seguir a estrutura de https://sol.sbc.org.br/index.php/sbie/article/view/22452/22276
\subsection{Definição das Perguntas de Pesquisa}

A pesquisa foi conduzida com o objetivo de compreender como os Modelos de Linguagem de Grande Escala (LLMs) são aplicados na análise de documentos jurídicos. Para isso, foram formuladas perguntas de pesquisa principais e secundárias, conforme apresentado na Tabela \ref{tab:perguntas}.

\begin{table}[h]

\centering
\caption{Perguntas de Pesquisa e Facetas}
\label{tab:perguntas}
\begin{tabular}{|c|p{10cm}|c|}

\hline
\textbf{ID} & \textbf{Pergunta de Pesquisa} & \textbf{Faceta} \\
\hline
RQ1 & Qual o modelo de linguagem utilizado? & Modelo \\
\hline
RQ2 & Quais as técnicas de engenharia de prompt utilizadas? & Técnica \\
\hline
RQ3 & Como a solução é avaliada? & Avaliação \\
\hline
RQ4 & Quais as métricas usadas na avaliação? & Métrica \\
\hline
RQ5 & Qual o desempenho dos LLMs na tarefa? & Desempenho \\
\hline
RQ6 & Qual o domínio de aplicação? & Domínio \\
\hline
\end{tabular}
\label{tab:perguntas_pesquisa}
\end{table}

Esta pesquisa busca compreender de forma ampla o impacto dos LLMs na análise de documentos jurídicos. As perguntas secundárias foram elaboradas para permitir uma avaliação detalhada dos aspectos técnicos, metodológicos e aplicacionais dos modelos de linguagem.

\subsection{Bases de Dados e Estratégia de Busca}

Para garantir um levantamento abrangente da literatura científica sobre o tema, a pesquisa foi realizada nas bases de dados SCOPUS e Springer, que são amplamente reconhecidas pela indexação de estudos relevantes na área de inteligência artificial e direito.

A estratégia de busca foi definida para recuperar artigos relacionados à aplicação de engenharia de \textit{prompts} e modelos de linguagem na análise de documentos jurídicos. A busca foi limitada ao período de 2020 a 2024 para garantir a inclusão de estudos recentes e relevantes. A query utilizada para a busca dos estudos foi:

( \textquote{Prompt Engineering} OR \textquote{Prompt Optimization} OR \textquote{Large Language Models} OR \textquote{LLMs} OR \textquote{generative artificial intelligence} ) AND ( \textquote{Legal Document} OR \textquote{Judicial Processes} OR \textquote{Legal Cases} OR \textquote{Court Documents} )

\subsection{Seleção de Estudos}
% falar que usou o chatgpt como suporte

Os estudos foram selecionados com base em critérios de inclusão e exclusão previamente estabelecidos, conforme apresentado na Tabela \ref{tab:criterios}. A seleção foi realizada em três etapas: (i) leitura dos títulos e resumos, (ii) análise do texto completo e (iii) avaliação final com base nos critérios de inclusão e exclusão.
Para garantir uma triagem eficiente e sistemática das publicações, utilizou-se o ChatGPT como suporte na organização e análise preliminar dos artigos. O modelo de linguagem foi empregado para identificar padrões nos dados, sugerir categorizações e auxiliar na extração de informações relevantes dos textos científicos, tornando o processo mais ágil e estruturado.
No entanto, todas as decisões finais sobre a inclusão ou exclusão dos estudos foram realizadas manualmente, garantindo rigor metodológico e alinhamento com os objetivos da pesquisa. Dessa forma, a ferramenta foi utilizada como um recurso complementar para otimizar a análise, sem comprometer a integridade dos resultados.

\begin{table}[h]
\centering
\caption{Critérios de Inclusão e Exclusão}
\label{tab:criterios}
\begin{tabular}{|c|p{12cm}|}
\hline
\textbf{Tipo} & \textbf{Critério} \\
\hline
\textbf{Inclusão} & IC1: O estudo aborda a aplicação de técnicas de engenharia de \textit{prompts} na análise de documentos jurídicos. \\
\hline
 & IC2: O estudo explora o uso de modelos de linguagem (LLMs) na interpretação de processos judiciais ou outros documentos legais. \\
\hline
\textbf{Exclusão} & EC1: O estudo está escrito em uma língua diferente de português ou inglês. \\
\hline
 & EC2: O estudo é um editorial, keynote, biografia, opinião, tutorial, workshop, resumo de relatório, poster, tese, dissertação, capítulo de livro ou painel. \\
\hline
 & EC3: O estudo não aborda a análise ou síntese de documentos jurídicos. \\
\hline
 & EC4: O estudo não utiliza técnicas de engenharia de \textit{prompts} ou Large Language Models (LLMs). \\
\hline
 & EC5: O foco do estudo é em outra área do direito que não envolve processos ou documentos jurídicos complexos em Língua Portuguesa. \\
\hline
\end{tabular}
\label{tab:criterios_inclusao_exclusao}
\end{table}

\subsection{Extração de Dados}

A extração de dados foi realizada de maneira sistemática para garantir uma análise comparativa entre os estudos selecionados. A Tabela 3 apresenta os principais elementos extraídos de cada estudo.

\begin{table}[h]
\centering
\caption{Elementos Extraídos dos Estudos}
\begin{tabular}{|l|p{8cm}|}
\hline
\textbf{Elemento} & \textbf{Descrição} \\
\hline
\textbf{Referência do estudo} & Autor, ano, título e fonte do estudo. \\
\hline
\textbf{Modelos de linguagem utilizados} & LLMs empregados na análise de documentos jurídicos. \\
\hline
\textbf{Técnicas de engenharia de prompt} & Métodos utilizados para otimizar a interação com os LLMs. \\
\hline
\textbf{Métodos de avaliação} & Procedimentos aplicados para validar a eficácia das soluções. \\
\hline
\textbf{Métricas de desempenho} & Indicadores utilizados para medir a qualidade das respostas. \\
\hline
\textbf{Principais achados e limitações} & Resultados obtidos e desafios identificados nos estudos. \\
\hline
\end{tabular}
\label{tab:extracao_dados}
\end{table}

\section{Resultados}

Na Tabela \ref{tab:resultados}, apresenta-se a revisão sistemática de artigos científicos sobre o uso de LLMs na área jurídica, explorando aspectos fundamentais como os modelos utilizados, as técnicas de engenharia de prompt, os métodos de avaliação das soluções, as métricas empregadas, o desempenho dos modelos e os domínios de aplicação. A partir dessa análise, é possível identificar tendências, desafios e avanços na aplicação dessas tecnologias para a compreensão e síntese de documentos jurídicos.

\input{tabelas/resultados}

O avanço dos LLMs tem proporcionado mudanças significativas no campo da inteligência artificial aplicada ao direito. Com a crescente digitalização dos processos jurídicos e a expansão do uso de documentos eletrônicos, os LLMs apresentam-se como uma ferramenta essencial para análise, classificação e geração de conteúdo legal. Neste contexto, diversos estudos têm sido conduzidos para avaliar o desempenho desses modelos na análise de documentos jurídicos. A partir das questões de pesquisa fundamentais (RQs), este trabalho busca explorar como os LLMs são empregados, suas limitações e potencialidades dentro do campo jurídico.

Modelos de Linguagem Utilizados (RQ1). Os modelos de linguagem empregados variam conforme a abordagem do estudo e o objetivo específico de cada análise. Muitos estudos utilizam modelos de uso genérico como GPT-3.5, GPT-4 e LLaMa 2 \cite{deroy2024applicability, wu2023precedentenhancedlegaljudgmentprediction, Venkatakrishnan2024sematic, deroy2023readypretrainedabstractivemodels, zhou2023boosting, Kang_2023, hijazi2024arablegaleval, chen2024general2, choi2024use, nunes2024out}. Alguns estudos optam por usar também por modelos especializados, como Legal-Pegasus e o Legal-LED \cite{deroy2023readypretrainedabstractivemodels,deroy2024applicability}. Enquanto outros optaram pelo BERT adaptados ao domínio jurídico \cite{zhou2023boosting,coelho2024information}. Os estudos demonstram, que a escolha do modelo influencia diretamente a capacidade do sistema de interpretar textos legais, pois modelos treinados em domínios específicos demonstram melhor compreensão dos conceitos jurídicos e maior precisão na classificação de documentos e extração de informações.

Técnicas de Engenharia de Prompt Utilizadas (RQ2). Para otimizar o desempenho dos modelos de linguagem, diferentes técnicas de engenharia de prompt são utilizadas. As estratégias incluem zero-shot prompting, no qual o modelo é solicitado a realizar uma tarefa sem exemplos prévios, e few-shot prompting, que fornece alguns exemplos para guiar a resposta do modelo. O Chain of Thought (CoT) \cite{wei2022chain}, por sua vez, incentiva o raciocínio passo a passo antes de fornecer uma resposta final. Outros métodos, como prompts adaptativos e reformulação dinâmica de consultas, também são utilizados para aprimorar a precisão das respostas, principalmente em contextos complexos como a extração de informações de pareceres jurídicos.

Avaliação da Solução (RQ3). A avaliação do desempenho dos modelos de linguagem no domínio jurídico ocorre por meio de diferentes abordagens. Algumas pesquisas utilizam métricas automáticas, como a similaridade entre resumos gerados e padrões de referência, enquanto outras adotam avaliação qualitativa conduzida por especialistas da área jurídica \cite{deroy2024applicability,wu2024knowledgeinfusedlegalwisdomnavigating}. Em estudos de classificação de documentos, por exemplo, os modelos são comparados com abordagens tradicionais, como aprendizado supervisionado e sistemas baseados em regras. Em determinados casos, os resultados são revisados por estudantes de direito e advogados para avaliar a coerência e precisão das respostas geradas.

Métricas Utilizadas na Avaliação (RQ4). A qualidade das saídas geradas pelos LLMs é medida através de um conjunto variado de métricas, dependendo da tarefa em questão. Para sumarização de textos, são amplamente empregadas métricas tradicionais para sumarização de texto, como ROUGE, METEOR e BLEU \cite{deroy2024applicability, deroy2023readypretrainedabstractivemodels, wu2024knowledgeinfusedlegalwisdomnavigating, hijazi2024arablegaleval, chen2024general2}, que avaliam a similaridade entre os resumos automáticos e os produzidos por humanos. No caso de classificação de textos e extração de entidades nomeadas, as métricas mais utilizadas são F1-score, Precisão e Recall \cite{coelho2024information, ghosh2023dalegenerativedataaugmentation, wu2023precedentenhancedlegaljudgmentprediction, hijazi2024arablegaleval, choi2024use}. Quando se trata de recuperação de informações, se observa o uso de métricas como MAP (Mean Average Precision) e NDCG (Normalized Discounted Cumulative Gain), que medem a relevância dos documentos retornados para uma consulta \cite{zhou2023boosting}.

Desempenho dos Modelos de Linguagem na Tarefa (RQ5). O desempenho dos LLMs varia significativamente de acordo com o modelo utilizado e a complexidade da tarefa. Em geral, modelos jurídicos especializados, como Legal-BERT e Sabiá, tendem a apresentar melhores resultados do que modelos genéricos como GPT-3.5 e GPT-4, especialmente em tarefas que envolvem terminologia técnica. No entanto, mesmo os modelos mais avançados enfrentam desafios relacionados às chamadas "alucinações" — respostas que parecem plausíveis, mas contêm informações erradas. Para mitigar esse problema, algumas abordagens combinam aprendizado supervisionado com interação humana para refinar os resultados.

Domínio de Aplicação (RQ6). Os modelos de linguagem são aplicados a diversas áreas do direito, variando de acordo com a necessidade específica do estudo. Alguns artigos se concentram na sumarização de decisões judiciais em sistemas legais da Índia e do Reino Unido \cite{deroy2024applicability}, enquanto outros exploram a recuperação de casos jurídicos no contexto chinês \cite{zhou2023boosting}. Há também aplicações voltadas para o reconhecimento de entidades nomeadas em textos legislativos brasileiros e a extração de informações de processos judiciais americanos. A versatilidade dos LLMs permite que sejam empregados em múltiplos contextos, desde a automação de tarefas rotineiras até análises complexas de cenários jurídicos.

\section{Conclusões}

O avanço dos Modelos de Linguagem de Grande Escala (LLMs) tem proporcionado impactos significativos na análise e síntese de documentos jurídicos. A crescente digitalização dos processos legais exige soluções cada vez mais eficazes para lidar com a complexidade e o volume crescente de informações. Nesse contexto, os LLMs se apresentam como ferramentas promissoras para otimizar a extração de informações, a sumarização de decisões judiciais e a automação de tarefas repetitivas no direito.

A revisão sistemática realizada neste estudo demonstrou que diferentes modelos de LLMs têm sido aplicados na área jurídica, desde os modelos generalistas, como GPT-4, Llama 2 e BERT, até aqueles mais especializados no contexto legal, como Legal-Pegasus e Sabiá. A escolha do modelo influencia diretamente o desempenho nas tarefas propostas, com os modelos treinados especificamente para o domínio jurídico apresentando maior precisão na classificação de textos, recuperação de jurisprudência e análise de contratos.
Outro aspecto essencial abordado foi a engenharia de prompts, que se mostra crucial para maximizar o desempenho dos LLMs. Técnicas como Zero-shot Learning, Few-shot Learning e Chain-of-Thought prompting foram amplamente utilizadas para aprimorar a precisão das respostas geradas pelos modelos. Os estudos analisados demonstraram que a formulação adequada de prompts pode reduzir significativamente as alucinações e viéses, tornando os modelos mais confiáveis para aplicações jurídicas.

A avaliação das soluções apresentou grande diversidade de metodologias. Alguns estudos adotaram métricas quantitativas, como ROUGE, METEOR, BLEU e F1-score, enquanto outros utilizaram avaliações qualitativas, com a participação de especialistas jurídicos para verificar a coerência e confiabilidade das respostas dos modelos. Os resultados indicam que, embora os LLMs tenham mostrado grande potencial na automação de tarefas jurídicas, ainda existem limitações que precisam ser superadas, como a necessidade de maior contextualização e a mitigação de erros sistemáticos.

No que diz respeito ao desempenho dos LLMs, observou-se que os modelos especializados na área jurídica tendem a superar os modelos generalistas em tarefas específicas. No entanto, ainda há desafios relacionados à interpretabilidade dos resultados e à dificuldade dos modelos em lidar com textos jurídicos altamente complexos e técnicos. Para que os LLMs sejam amplamente adotados no setor jurídico, é fundamental investir em treinamentos mais refinados e na criação de datasets jurídicos robustos que possam aprimorar a adaptação desses modelos ao domínio legal.

Por fim, a análise do domínio de aplicação dos LLMs revelou que essas tecnologias estão sendo empregadas em diversas frentes do direito, incluindo sumarização de decisões judiciais, extração de entidades nomeadas, recuperação de informações em bancos de dados jurídicos e predição de resultados de processos. No entanto, sua aplicação em larga escala ainda exige estudos mais aprofundados, especialmente para garantir a transparência, confiabilidade e conformidade com as normativas jurídicas vigentes.

Em conclusão, este estudo destacou que os LLMs representam uma ferramenta poderosa para transformar o setor jurídico, trazendo eficiência e acessibilidade ao processamento de informações legais. No entanto, seu uso ainda requer avanços na engenharia de prompts, no controle de viés e na confiabilidade dos modelos. Pesquisas futuras devem explorar estratégias de ajuste fino (fine-tuning) mais eficazes, além do desenvolvimento de modelos jurídicos mais especializados para diferentes sistemas legais e línguas. Dessa forma, será possível maximizar o impacto positivo dessas tecnologias no campo do direito, garantindo que os LLMs sejam aplicados de forma ética, precisa e eficiente

\bibliographystyle{sbc}
\bibliography{sbc-template}

/

\end{document}

%% file: tabelas/resultados.tex
% \begin{adjustbox}{angle=90}
\begingroup

\setlength{\tabcolsep}{3pt} % Default value: 6pt
\renewcommand{\arraystretch}{1} % Default value: 1

\begin{landscape}

\begin{table}[]
\caption{Resultados}
\footnotesize
\label{tab:resultados}
\begin{longtable}{|p{2.8cm}|p{3.3cm}|p{3.3cm}|p{3.3cm}|p{3.3cm}|p{3.3cm}|p{3.3cm}|}
\hline
\textbf{Artigo}                                                                                                                         & \textbf{RQ1: Modelo}                         & \textbf{RQ2: Prompt}                                              & \textbf{RQ3: Avaliação}                                              & \textbf{RQ4: Métricas}                        & \textbf{RQ5: Desempenho}                                             & \textbf{RQ6: Domínio}                                                  \\ \hline
\cite{deroy2024applicability}                                      & GPT-4 e 3.5, LLama-2, Legal-Pegasus, Legal-LED & TL;DR  & Métricas automáticas e avaliação com estudantes de Direito & Rouge, METEOR, BERTScore, SummaC, NumPrec & LLMs superaram métodos extrativos  & Sumarização de decisões judiciais do Reino Unido e Índia                   \\ \hline
\cite{ghosh2023dalegenerativedataaugmentation}                                                                            & BART (Encoder-Decoder)                                     & Mascaramento seletivo de spans                         & Testado em 13 datasets de 6 tarefas & F1, Precisão, Diversidade de Tokens               & Melhorou o desempenho em 1\%-50\%       &  Aumento de dados para NLP jurídico                              \\ \hline
\cite{wu2023precedentenhancedlegaljudgmentprediction}                                                     & ChatGPT (GPT-3.5)                                          & Reorganização de fatos jurídicos        & Experimentos com CAIL2018 e CJO22                                      & Acurácia, Precisão, Recall, F1  & Proposta superou baseline                                  & Predição de julgamentos na China                                           \\ \hline
[Venkatakrishnan et al. 2024]                                                   & GPT-3.5, Mistral 8*7B                                      & Construção de grafos         & Avaliação qualitativa dos grafos                        & Não especificado                                  & Geração de grafos com alta precisão                                  & Análise de dados de imigração                                              \\ \hline
\cite{deroy2023readypretrainedabstractivemodels}                                            & Text-Davinci-003, GPT-3.5, modelos ajustados  & Tl;Dr, divisão em blocos de 1024 palavras                              & Similaridade entre resumos gerados e padrões                           & ROUGE, METEOR, BLEU, SummaC, NEPrec, NumPrec      & Modelos jurídicos específicos tiveram melhor desempenho              & Sumarização de decisões judiciais indianas                                 \\ \hline
\cite{wu2024knowledgeinfusedlegalwisdomnavigating}        & D3LM fine-tune do LLaMa-2                         & Perguntas diagnósticas       & avaliação automática e humana                 & ROUGE, BLEU, fluência, precisão                   & Superou GPT-4 em métricas específicas                                & Análise de casos criminais nos EUA                                         \\ \hline
\cite{zhou2023boosting}                                                      & GPT-3.5-turbo, BERT                         & Sumarização extrativa                     & Experimentos com modelos de busca                       & P@5, P@10, MAP, NDCG                              & Melhor que baselines                                & Recuperação de casos jurídicos                                    \\ \hline
\cite{Kang_2023}                                          & ChatGPT                                                    & Decomposição de perguntas, in-context learning, CoT & Avaliação por estudantes e especialistas jurídicos                     & Precisão, recall, F1-score, Cohen’s Kappa         & Desempenho misto, F1 médio de 0.49                                   & Direito contratual e familiar (Malásia e Austrália)                        \\ \hline
\cite{prasad2024goharian}                     & GPT-Neo, GPT-J, LEGAL-BERT, InLegalBERT                    & Segmentação em chunks, embeddings de última camada                     & Testes em datasets jurídicos                                           & Micro-F1, Macro-F1, acurácia                      & MESc teve desempenho superior aos métodos anteriores                 & Classificação de documentos jurídicos (Índia, UE, EUA)                     \\ \hline
\cite{nunes2024out}                                   & Sabiá (Llama 7B ajustado para português)                   & In-Context Learning com heurísticas K similares                        & Testes em corpora brasileiros (LeNER-Br, UlyssesNER-Br)                & F1-micro, precisão, recall                        & Melhor desempenho com K similares, F1 de 51\%                        & NER em textos jurídicos brasileiros \\ \hline
\cite{hijazi2024arablegaleval}                                       & GPT-4, Jais, Llama 3, Claude-3-opus                        & In-Context Learning, Chain of Thought, Few-shot e Zero-shot            & Testes em MCQs, Q\&A e traduções                                       & F1-score, ROUGE, métrica de similaridade GPT-4    & GPT-4 foi o melhor geral, Jais destacou-se no direito saudita        & Direito saudita e árabe                                                    \\ \hline
\cite{chen2024general2}                                                                                        & Qwen-14B, LLaMA, GPT-3.5, GPT-4                            & Ajuste fino em duas etapas, R-Drop                                     & Testes com corpus de e-commerce                                        & SacreBLEU, ROUGE              & G2ST superou modelos de última geração                               & Tradução de textos de e-commerce                                           \\ \hline
\cite{coelho2024information}                                                  & ChatGPT, Doc2Vec, C-LSTM, BERT     & Prompts específicos com justificativa                          & Testes em pareceres jurídicos brasileiros                              & Precisão, recall, F1-score, acurácia, RMSE        & Desempenho competitivo & Extração de informações jurídicas                               \\ \hline
\cite{choi2024use}                                                                            & GPT-4 e 3.5 Turbo, RandomForest, SVM                   & Zero-shot, Few-shot, Chain-of-Thought                                  & Comparação com  humanos                                   & Acurácia, precisão, recall, F1-score              & GPT-4 comparável a humanos (F1=0.89)                                 & Classificação de opiniões         \\ \hline
% \end{tabular}
\end{longtable}
\end{table}
\end{landscape}

% \end{adjustbox}
\endgroup